# Bayesian Neural Networks: Essentials

Daniel T. Chang (张遵)

*IBM (Retired)* dtchang43@gmail.com

**Abstract:** Bayesian neural networks utilize probabilistic layers that capture uncertainty over weights and activations, and are trained using Bayesian inference. Since these probabilistic layers are designed to be drop-in replacement of their deterministic counter parts, Bayesian neural networks provide a direct and natural way to extend conventional deep neural networks to support probabilistic deep learning. However, it is nontrivial to understand, design and train Bayesian neural networks due to their complexities. We discuss the essentials of Bayesian neural networks including duality (deep neural networks, probabilistic models), approximate Bayesian inference, Bayesian priors, Bayesian posteriors, and deep variational learning. We use TensorFlow Probability APIs and code examples for illustration. The main problem with Bayesian neural networks is that the architecture of deep neural networks makes it quite redundant, and costly, to account for uncertainty for a large number of successive layers. Hybrid Bayesian neural networks, which use few probabilistic layers judicially positioned in the networks, provide a practical solution.

## 1 Introduction and Overview

*Probabilistic deep learning* [1] is deep learning that accounts for uncertainty, both model uncertainty and data uncertainty. It is based on the use of *probabilistic models* and *deep neural networks*. A widely studied approach to probabilistic deep learning is Bayesian neural networks [3].

*Bayesian neural networks* [1-2] utilize *probabilistic layers* that capture uncertainty over weights and activations, and are trained using *Bayesian inference* [4]. Since these probabilistic layers are designed to be drop-in replacement of their deterministic counter parts, Bayesian neural networks provide a direct and natural way to extend conventional deep neural networks to support probabilistic deep learning. However, it is nontrivial to understand, design and train Bayesian neural networks, even for simple applications [2], due to their complexities.

We discuss the essentials of Bayesian neural networks including duality, approximate Bayesian inference, Bayesian priors, Bayesian posteriors, and deep variational learning. Bayesian neural networks are *deep neural networks*; they are as well *probabilistic models*. The *duality* makes Bayesian neural networks inherently more complex to understand, design and train than either deep neural networks or probabilistic models alone.

Bayesian neural networks are trained using Bayesian inference. Unfortunately, computing the Bayesian posterior, and moreover sampling from it, is usually intractable. *Approximate Bayesian inference* [3, 6] methods, therefore, are generally used, including *Markov Chain Monte Carlo* [3, 5] and *Variational Inference* [3, 8].

*Bayesian priors p(θ)* [4, 7] let us encode our beliefs about the parameters **θ** of a Bayesian neural network before seeing any data. For basic Bayesian neural network architectures, a standard procedure [3] is to use an isotropic *Gaussian prior*, which is favored for its mathematical properties and the simple formulation of its log.

*Bayesian posteriors p(θ | D)*, given observed data D, are the basis for making prediction in Bayesian neural networks, using Bayesian posterior predictive. However, since computing the Bayesian posterior is intractable, in general *approximate posteriors* [8-10] are used, and usually they are obtained through *variational inference*. For Bayesian neural networks with Gaussian priors, it is natural to use *Gaussian approximate posteriors* [8-9].

*Deep variational learning* is the adaptation of variation inference to deep neural networks, which conventionally use *stochastic gradient descent* and *backpropagation* for training. The main problem in doing so is that stochasticity in variational inference stops backpropagation from functioning for the hidden nodes in a network. *Bayes By Backprop (BBB)* [3, 11] is a practical implementation of stochastic variational inference combined with a reparametrization trick to ensure backpropagation works as usual.

The main problem [3] about deep variational learning for Bayesian neural networks is that the architecture of deep neural networks makes it quite redundant, and costly, to account for uncertainty for a large number of successive layers. Hybrid Bayesian neural networks, which use few probabilistic layers judicially positioned in the networks, provide a practical solution.

We use TensorFlow Probability [13] APIs and code examples, where applicable, for illustration.

## 2 Duality of Bayesian Neural Networks

Bayesian neural networks (BNNs) are *deep neural networks*; they are as well *probabilistic models*. The duality makes BNNs inherently more complex to understand, design and train than either deep neural networks or probabilistic models alone.

## 2.1 BNNs as Deep Neural Networks

As *deep neural networks*, in the simplest architecture, BNNs can be represented as follow [3], with **x** as input variables, **y** as output variables (predictions) and **h** as hidden variables:

- $\mathbf{h}_0 = \mathbf{x}$
- $\mathbf{h}_i = \mathcal{A}_i(\mathbf{W}_i \mathbf{h}_{i-1} + \mathbf{b}_i) \quad \forall i \in [1, n]$
- $\mathbf{y} = \mathbf{h}_n$

where at each layer **h** is represented as a linear transformation of the previous one, followed by a nonlinear *activation function* $\mathcal{A}$. This means that a BNN represents a set of functions isomorphic to the set of *parameter* $\theta$, which represent all the *weights W* and *biases b* of the network.

Deep neural networks are trained using stochastic gradient descent and back propagation. *Stochastic gradient descent* is an iterative method for optimizing an *objective function* (e.g., loss function) with suitable smoothness properties (e.g. differentiable). *Backpropagation* works by computing the gradient of the loss function with respect to each weight (and bias) by the *chain rule*, computing the gradient one layer at a time, iterating backward from the last layer. Both stochastic gradient descent and back propagation have been well discussed in the deep learning literature.

In *TensorFlow Probability*, BNNs are supported using probabilistic layers, *tfp.layers*, e.g., Convolution2DFlipout and DenseFlipout, which are drop-in replacements of their conventional counterparts Convolution2D and Dense, respectively. The following code example [13] shows a convolutional BNN model using 3 Convolution2DFlipout layers and 2 DenseFlipout layers. Note that the model outputs point estimates. Also, the code example for defining kl_divergence_function is shown later in 3.2 Variational Inference.

```
model = tf.keras.models.Sequential([
    tfp.layers.Convolution2DFlipout(
        6, kernel_size=5, padding='SAME',
        kernel_divergence_fn=kl_divergence_function,
        activation=tf.nn.relu),
    tf.keras.layers.MaxPooling2D(
        pool_size=[2, 2], strides=[2, 2],
        padding='SAME'),
    tfp.layers.Convolution2DFlipout(
        16, kernel_size=5, padding='SAME',
        kernel_divergence_fn=kl_divergence_function,
        activation=tf.nn.relu),
    tf.keras.layers.MaxPooling2D(
        pool_size=[2, 2], strides=[2, 2],
        padding='SAME'),
    tfp.layers.Convolution2DFlipout(
        120, kernel_size=5, padding='SAME',
        kernel_divergence_fn=kl_divergence_function,
        activation=tf.nn.relu),
    tf.keras.layers.Flatten(),
    tfp.layers.DenseFlipout(
        84, kernel_divergence_fn=kl_divergence_function,
        activation=tf.nn.relu),
    tfp.layers.DenseFlipout(
        NUM_CLASSES, kernel_divergence_fn=kl_divergence_function,
        activation=tf.nn.softmax)
])
```

## 2.2 BNNs as Probabilistic Models

*Probabilistic models* are trained from data using *Bayesian inference* [4]. The main idea is to infer a *posterior distribution (Bayesian posterior)* over the parameters $\boldsymbol{\theta}$ of a probabilistic model given some observed data D using Bayes theorem as:

$$p(\boldsymbol{\theta} \mid D) = p(D \mid \boldsymbol{\theta})p(\boldsymbol{\theta}) / p(D) = p(D \mid \boldsymbol{\theta})p(\boldsymbol{\theta}) / \int p(D \mid \boldsymbol{\theta})p(\boldsymbol{\theta})d\boldsymbol{\theta},$$

where $p(D \mid \boldsymbol{\theta})$ is the *likelihood*, $p(D)$ is the *marginal likelihood* (*evidence*), and $p(\boldsymbol{\theta})$ is the *prior (Bayesian prior)*.

The Bayesian posterior can be used to model new unseen data D* using the *posterior predictive*:

$$p(D^* \mid D) = \int p(D^* \mid \boldsymbol{\theta})p(\boldsymbol{\theta} \mid D)d\boldsymbol{\theta},$$

which is also called the *Bayesian model average* because it averages the predictions of all plausible models weighted by their posterior probability.

As probabilistic models, BNNs can be represented as follow [3]:

- $\theta \sim p(\theta)$
- $y = BNN_\theta(x) + \epsilon$

where $\theta$ represents BNN parameters and $\epsilon$ represents random noise.

To design a BNN, the first step is to choose a neural network architecture, e.g., a convolutional neural network. Then, one has to choose a probability model: a *Bayesian prior* over the possible model parameters $p(\theta)$ and a *prior confidence* in the predictive power of the model $p(y \mid x, \theta)$, i.e., $BNN_\theta(x)$. For supervised learning, the *Bayesian posterior* can be written as:

$$p(\theta \mid D) = p(D_y \mid D_x, \theta)p(\theta) / \int p(D_y \mid D_x, \theta)p(\theta)d\theta,$$

where D is the training set, $D_x$ the training features, and $D_y$ the training labels. And the *posterior predictive* can be computed as:

$$p(y \mid x, D) = \int p(y \mid x, \theta)p(\theta \mid D)d\theta.$$

In practice, the distribution $p(y \mid x, \theta)$ is sampled indirectly using $BNN_\theta(x)$, and $\theta$ is sampled from an approximate $p(\theta \mid D)$, i.e., an *approximate posterior*, as discussed later.

The following *TensorFlow Probability* code example [2] shows a *probabilistic BNN model* that outputs a normal (`IndependentNormal`) distribution with learnable mean and variance parameters. Note that the code examples for defining the prior ('prior') and the approximate posterior ('posterior') are shown later in 4.1 Gaussian Priors and 5.1 Gaussian Approximate Posteriors, respectively.

```python
def create_probablistic_bnn_model(train_size):
    inputs = create_model_inputs()
    features = keras.layers.concatenate(list(inputs.values()))
    features = layers.BatchNormalization()(features)

    # Create hidden layers with weight uncertainty using the
    # DenseVariational layer.
    for units in hidden_units:
        features = tfp.layers.DenseVariational(
            units=units,
            make_prior_fn=prior,
            make_posterior_fn=posterior,
            kl_weight=1 / train_size,
            activation="sigmoid",
        )(features)

    # Create a probabilistic output (Normal distribution).
    distribution_params = layers.Dense(units=2)(features)
    outputs = tfp.layers.IndependentNormal(1)(distribution_params)

    model = keras.Model(inputs=inputs, outputs=outputs)
    return model
```

## 3 Approximate Bayesian Inference

Computing the Bayesian posterior, and moreover sampling from it, is usually intractable, since computing the evidence (integrals) is hard. *Approximate Bayesian inference* methods are generally used, including:

1. *Markov Chain Monte Carlo*: Approximates integrals via sampling.
2. *Variational Inference*: Approximates integrals via optimization.

### 3.1 Markov Chain Monte Carlo

*Markov Chain Monte Carlo (MCMC)* methods [3] construct a *Markov chain*, a sequence of random samples $S_i$ of the Bayesian posterior $p(\theta \mid D)$, which probabilistically depend only on the previous sample $S_{i-1}$, such that the elements of the sequence eventually are distributed following $p(\theta \mid D)$. The class of MCMC methods which are the most relevant for Bayesian posteriors are the *Metropolis-Hastings algorithms*, which do not require knowledge about the exact distribution to sample from, only a function that is proportional to that distribution. This is the case of Bayesian posteriors which are usually quite easy to compute except for the evidence term.

*Hamiltonian Monte Carlo (HMC)* algorithms [5] are examples of Metropolis-Hastings algorithms, and have been a popular choice of inference due to their ability to suppress random-walk behavior through the use of first-order gradient information. However, they still suffer from the presence of autocorrelations in the generated samples, which results in the high variance of their estimations.

In deep learning, the large volume of data and the huge size of models used in practice make MCMC methods *too slow to be used* [6]: first, each iteration of the algorithm requires accessing all the data, then the number of iterations required to reach convergence explodes when the dimension is large. Therefore, we will not discuss MCMC methods in detail or its usage in BNNs; we will focus instead on *variational inference* and its usage in BNNs in subsequent discussions.

In *TensorFlow Probability*, MCMC methods are supported using *tfp.mcmc*, e.g., MetropolisHastings and HamiltonianMonteCarlo.

### 3.2 Variational Inference

*Variational inference* [3, 8] learns a *variational distribution* $q_\phi(\theta)$, parameterized by a set of parameters $\phi$, to approximate the Bayesian posterior $p(\theta \mid D)$. The values of the parameters $\phi$ is inferred such that the variational distribution is as close as possible to the Bayesian posterior.

The measure of closeness commonly used is the *(reverse) KL-divergence*, which is a measure of closeness between probability distributions, and is a function of $\phi$:

$$D_{KL}(q_\phi \| p) = \int q_\phi(\theta) \log(q_\phi(\theta) / p(\theta \mid D)) d\theta.$$

Minimizing $D_{KL}(q_\phi \| p)$ is equivalent to maximizing the *evidence lower bound (ELBO)* [3]:

$$\text{ELBO} = \int q_\phi(\theta) \log(p(\theta, D) / q_\phi(\theta)) d\theta = \log(p(D)) - D_{KL}(q_\phi \| p).$$

In *TensorFlow Probability*, variational inference is supported using *tfp.distributions.kl_divergence* and *tfp.vi*, the latter containing methods and objectives for variational inference. The following code example [13] shows the usage of *KL divergence* (tfp.distributions.kl_divergence). Using Lambda function, it is passed as input to the 'kernel_divergence_fn' on flipout layers. The Keras API automatically adds the KL divergence to the conventional loss function, e.g., cross entropy loss, effectively calculating the (negated) Evidence Lower Bound Loss (ELBO). Note that the code example for the full model has been shown in 2.2 BNNs as Probabilistic Models.

```
kl_divergence_function = (lambda q, p, _: tfd.kl_divergence(q, p) /
                         tf.cast(NUM_TRAIN_EXAMPLES, dtype=tf.float32))

model = tf.keras.models.Sequential([
    tfp.layers.Convolution2DFlipout(
        6, kernel_size=5, padding='SAME',
        kernel_divergence_fn=kl_divergence_function,
        activation=tf.nn.relu),
    …
    tfp.layers.DenseFlipout(
        NUM_CLASSES, kernel_divergence_fn=kl_divergence_function,
        activation=tf.nn.softmax)
])
```

## 4 Bayesian Priors

*Bayesian priors p(θ)* let us encode our beliefs about the parameters **θ** before seeing any data. There is no universally preferred prior: each probabilistic model / inference task is potentially endowed with its own optimal prior [4].

Unfortunately, the *weights* have no intuitive interpretation. Further, it is not really clear how models with a very large number of parameters and a nontrivial architecture like deep neural networks will generalize for a given parametrization [3]. Seemingly sensible priors can induce unintended artifacts in the Bayesian posteriors, and the details over the priors in weight space can have a relatively minor effect on prediction performance.

### 4.1 Gaussian Priors

For basic BNN architectures, such as Bayesian regression, a standard procedure [3] is to use an isotropic *Gaussian prior* with mean **0** and diagonal covariance $\sigma^2 \mathbf{I}$ on the parameters **θ** of the network:

$$p(\mathbf{\theta}) = \mathcal{N}(0, \sigma^2 \mathbf{I}).$$

A Gaussian prior is favored for its mathematical properties and the simple formulation of its log.

Gaussian priors are *uninformed*. However, there is no strong evidence that Gaussian priors are particularly bad. In fact, preliminary studies on small networks and simple problems did not find conclusive evidence for the misspecification of Gaussian priors [4].

The following *TensorFlow Probability* code example [2] shows the definition of a Gaussian prior. Note that the code example for setting the prior in a probabilistic BNN model has been shown in 2.2 BNNs as Probabilistic Models.

```
# Define the Gaussian prior of mean=0 and stddev=1.

def prior(kernel_size, bias_size, dtype=None):
    n = kernel_size + bias_size
    prior_model = keras.Sequential(
        [
            tfp.layers.DistributionLambda(
                lambda t: tfp.distributions.MultivariateNormalDiag(
                    loc=tf.zeros(n), scale_diag=tf.ones(n)
                )
            )
        ]
    )
    return prior_model
```

## 4.2 Induced Priors at the Unit Level

BNNs have *heavy-tailed deep units* [7] as found by a study of hidden units prior distributions in BNNs with *Gaussian priors*. The result establishes that the *induced (by forward propagation) prior distribution* on the units before and after activation becomes increasingly heavy-tailed with the depth of the layer. In particular, first layer units are *Gaussian*, second layer units are *sub-exponential*, and units in deeper layers are characterized by *sub-Weibull* distributions.

## 5 Bayesian Posteriors

*Bayesian posteriors* are the basis for making prediction in BNNs, using Bayesian posterior predictive. However, computing the Bayesian posterior, and moreover sampling from it, is usually intractable, since computing the evidence (integrals) is hard. Therefore, in general, *approximate posteriors* are used, and usually they are obtained through *variational inference*, as discussed previously.

## 5.1 Gaussian Approximate Posteriors

For BNNs with Gaussian priors, it is natural to use an *Gaussian approximate posterior* for $q_\phi(\boldsymbol{\theta})$, which is a fully factorized Gaussian distribution $\mathcal{N}(\mu_i, \sigma_i^2)$ over the parameter $\boldsymbol{\theta}_i$ assuming $\boldsymbol{\theta} = (\boldsymbol{\theta}_1, ..., \boldsymbol{\theta}_N)$. Variational inference using a Gaussian approximate posterior is exemplary of *mean-field variational inference (MFVI)* [8] and synonymous with MFVI. MFVI utilizes the mean-field approximation to simplify variational inference, which partitions $q_\phi(\boldsymbol{\theta})$ into factorized parts:

$$q_\phi(\boldsymbol{\theta}) = \prod_{i=1}^{N} q_\phi(\boldsymbol{\theta}_i).$$

In the case of a unit multivariate *Gaussian prior* and a *Gaussian approximate posterior*, the negative *evidence lower bound (ELBO)* is simplified to [10]:

$$\text{ELBO} = \Sigma_i((\mu_i^2 + \sigma_i^2)/2) - \Sigma_i \log(\sigma_i) - \mathbb{E}_{\boldsymbol{\theta} \sim q_\phi(\boldsymbol{\theta})}(\log(p(\mathbf{y} \mid \mathbf{x}, \boldsymbol{\theta}))),$$

with the first term being prior-posterior cross-entropy, the second approximate posterior entropy, and the third likelihood.

MFVI fails to give calibrated uncertainty estimates *in between separated regions of observations* [9]. This can lead to catastrophically overconfident predictions when testing on out-of-distribution data.

The following *TensorFlow Probability* code example [2] shows the definition of a Gaussian approximate posterior. Note that the code example for setting the approximate posterior in a probabilistic BNN model has been shown in 2.2 BNNs as Probabilistic Models.

```
# Define approximate posterior as multivariate Gaussian.
# Note that the learnable parameters for this distribution are the means,
# variances, and covariances.

def posterior(kernel_size, bias_size, dtype=None):
    n = kernel_size + bias_size
    posterior_model = keras.Sequential(
        [
            tfp.layers.VariableLayer(
                tfp.layers.MultivariateNormalTriL.params_size(n),
                dtype=dtype
            ),
            tfp.layers.MultivariateNormalTriL(n),
        ]
    )
    return posterior_model
```

## 5.2 Radial Approximate Posteriors

Typical samples from the *Gaussian approximate posterior* are unrepresentative of the most-probable weights, and this problem gets worse for larger networks [10]. This is due to the fact that probability distribution in a multivariate Gaussian distribution is clustered in a narrow *'soap-bubble' far from the mean*. This leads to *exploding gradient variance* whenever the posterior becomes broad, and prevents MFVI from actually fitting to the loss.

The *Radial approximate posterior* [10] defines a distribution, without a 'soap-bubble', in a *hyperspherical space* corresponding to each layer, and then transforms this distribution into the coordinate system of the weights. The typical samples from this distribution tend to come from areas of high probability density, more reflective of the mean. This helps training by *reducing gradient variance*.

In the *hyperspherical coordinate system*, the first dimension is the *radius* and the remaining dimensions are *angles*. The Radial approximate posterior is defined as:

- Radius dimension: $r = |r^*|$ for $r^* \sim \mathcal{N}(0, 1)$.

- Angles dimensions: uniform distribution over the hypersphere—all directions equally likely.

A critical property of the Radial approximate posterior is that it is easy to sample this distribution in the weight-space coordinate system. Instead of sampling the distribution directly, the local reparametrization trick is used and the noise distribution is sampled:

$$\boldsymbol{\theta} := \boldsymbol{\mu} + \boldsymbol{\sigma} \odot (\boldsymbol{\epsilon}_{MFVI} / \|\boldsymbol{\epsilon}_{MFVI}\|) * r$$

where $\boldsymbol{\epsilon}_{MFVI} \sim \mathcal{N}(0, \mathbf{I})$. As such, sampling from the Radial approximate posterior is nearly as cheap as sampling from the Gaussian approximate posterior. The only extra steps are to normalize the noise, and multiply by a scalar Gaussian random variable r.

## 6 Deep Variational Learning

Variation inference needs to be adapted to deep neural networks, which conventionally use *stochastic gradient descent* and *backpropagation* for training. We call the adaption *deep variational learning*.

### 6.1 Bayes By Backprop

The most popular optimization method for variational inference in deep neural networks is *stochastic variational inference* [3], which is *stochastic gradient descent* applied to variational inference. The main problem is that stochasticity stops backpropagation from functioning for the hidden nodes in a network.

*Bayes By Backprop (BBB)* [3, 11] is a practical implementation of stochastic variational inference combined with a reparametrization trick to ensure backpropagation works as usual. The idea of *reparametrization* is to use a random variable $\epsilon$, as a source of non-variational noise, such that

- $\epsilon \sim q(\epsilon)$
- $\boldsymbol{\theta} = t(\epsilon, \boldsymbol{\phi})$ follows $q_{\boldsymbol{\phi}}(\boldsymbol{\theta})$ and t is deterministic
- $q_{\boldsymbol{\phi}}(\boldsymbol{\theta})d\boldsymbol{\theta} = q(\epsilon)d\epsilon$

$\epsilon$ is sampled but can be considered as a constant with regard to other variables. With all other transformations being non-stochastic, *backpropagation* works as usual for the variational parameters $\boldsymbol{\phi}$.

The *BBB algorithm* [3] consists of the following steps, for each iteration:

1. Draw $\epsilon \sim q(\epsilon)$
2. $\theta = t(\epsilon, \phi)$
3. $f(\theta, \phi) = \log(q_\phi(\theta)) - \log(p(D_y | D_x, \theta)p(\theta))$
4. $\Delta_\phi f = \text{backprop}_\phi(f)$
5. $\phi = \phi - \alpha \Delta_\phi f$

The function f corresponds to an estimate of the *ELBO* from a single sample.

For *Gaussian approximate posteriors*, $\mathcal{N}(\mu, \sigma^2 I)$, we have specifically [11]:

- $\sigma = \log(1 + \exp(\rho))$ so that $\sigma$ is always non-negative
- $\epsilon \sim \mathcal{N}(0, I)$
- $\theta = t(\epsilon, \phi) = \mu + \log(1 + \exp(\rho)) \odot \epsilon$
- $\phi = (\mu, \rho)$
- $\mu = \mu - \alpha \Delta_\mu f$
- $\rho = \rho - \alpha \Delta_\rho f$

To learn both the *mean μ* and the *standard deviation σ* we must simply calculate the usual gradients found by backpropagation, as in Step 4 of the BBB algorithm, and then scale and shift them as in Step 5.

BBB trains an *ensemble of networks*, where each network has its weights drawn from a *shared, learnt probability distribution*. It typically only *doubles the number of parameters* yet trains an infinite ensemble of networks.

## 6.2 Hybrid BNNs

The main problem [3] about deep variational learning for BNNs is that the architecture of deep neural networks makes it quite redundant, and costly, to account for uncertainty for a large number of successive layers. A recent study [12] found that there is limited value in adding multiple uncertainty layers to deep classifiers.

The suggestion [3] is to use few probabilistic layers positioned at the end of the networks. Training only few probabilistic layers can drastically simplify the learning procedure. It removes many conception and training problems, with regard to uncertainty, while still being able to give meaningful results from a Bayesian neural network perspective. It can be interpreted as learning partly a *conventional neural network* and partly a *shallow BNN*, i.e., a *hybrid BNN*.

A hybrid BNN, with L number of layers, consists of *L-P deterministic layers* followed by *P probabilistic layers*, where P being one or a small number. The deterministic layers are used for *representation learning* to learn the task-specific representation (high-level features); the probabilistic layers are used for *prediction generation and uncertainty estimation*, based on the task-specific representation (high-level features). The "separation" of representation learning followed by prediction generation is commonly seen in conventional deep neural networks which use *last few dense layers* for prediction generation. Therefore, it is straightforward and natural to convert these conventional deep neural networks to hybrid BNNs by simply converting the last few dense layers to *probabilistic dense layers*.

A particular implementation of a hybrid BNN (with P = 1) is discussed in [12]. For classification tasks, the hybrid BNN is *first trained as a conventional deep neural network* to convergence using the cross entropy loss and a conventional optimizer. Then a *two-stage procedure* is used. First, the *last layer features* $\mathbf{z}$ from the inputs $\mathbf{x}$ is computed by making a *forward pass through the trained network*. This is done for all points in D, and produces a new training dataset R = $\{\mathbf{z}_i, \mathbf{y}_i\}_{i=1}^{N}$. Second, prediction and uncertainty estimation are carried out on R, applied to the last layer, via *an approximate Bayesian inference algorithm*. The last layer of the network in this case is a dense layer with a softmax activation. Note that the pretrained network and the two-stage procedure may make the *Bayesian inference* not hold. They also make hybrid BNNs cumbersome to use.

Ideally, no pretraining is required and a single-stage seamless training is all that needed. This appears to be doable in TensorFlow Probability, as discussed below.

## TensorFlow Probability

In *TensorFlow Probability*, deterministic layers *(keras.layers)* and probabilistic layers *(tfp.layers)* can be used together. However, we should note that there is no discussion of usage patterns or usage constraints in the documentation or elsewhere. We experiment with two different combinations of deterministic layers and probabilistic layers for a regression task with a fully connected network. For Case 1, with example code shown in 2.2 BNNs as Probabilistic Models, the hybrid BNN consists of two probabilistic ('DenseVariational') hidden layers followed by a deterministic ('Dense') output layer. In contrast, for Case 2 with example code shown below, the hybrid BNN consists of two deterministic ('Dense') hidden layers followed by a probabilistic ('DenseVariational') output layer. Not that the code examples for defining the prior ('prior') and approximate posterior ('posterior') has been shown in 4.1 Gaussian Priors and 5.1 Gaussian Approximate Posteriors, respectively.

```
def create_probablistic_hbnn_model(train_size):
    inputs = create_model_inputs()
    features = keras.layers.concatenate(list(inputs.values()))
    features = layers.BatchNormalization()(features)

    # Create hidden layers with deterministic Dense layer.
    for units in hidden_units:
        features = layers.Dense(units, activation="sigmoid")(features)

    # Create a probabilisticâ output (Normal distribution), and use the
    # `DenseVariational` layer with weight uncertainty,
    # to produce the parameters of the distribution.
    # We set units=2 to learn both the mean and the variance of the Normal
    # distribution.

    distribution_params = tfp.layers.DenseVariational(
        units=2,
        make_prior_fn=prior,
        make_posterior_fn=posterior,
        kl_weight=1 / train_size,
    )(features)
    outputs = tfp.layers.IndependentNormal(1)(distribution_params)

    model = keras.Model(inputs=inputs, outputs=outputs)
    return model
```

Case 2 generates a slightly better result than Case 1 with better predictions (means) and smaller variances. The result of Case 2, with two deterministic hidden layers followed by a probabilistic output layer, is shown below for eight randomly selected samples,:

```
Prediction mean: 5.96, stddev: 0.69, 95% CI: [7.32 - 4.6]  - Actual: 6.0
Prediction mean: 5.83, stddev: 0.71, 95% CI: [7.24 - 4.43] - Actual: 5.0
Prediction mean: 5.81, stddev: 0.7,  95% CI: [7.17 - 4.44] - Actual: 6.0
Prediction mean: 6.14, stddev: 0.74, 95% CI: [7.59 - 4.69] - Actual: 6.0
Prediction mean: 6.81, stddev: 0.74, 95% CI: [8.26 - 5.35] - Actual: 8.0
Prediction mean: 5.46, stddev: 0.72, 95% CI: [6.86 - 4.05] - Actual: 5.0
Prediction mean: 5.4,  stddev: 0.72, 95% CI: [6.81 - 4.0]  - Actual: 6.0
Prediction mean: 5.12, stddev: 0.73, 95% CI: [6.56 - 3.69] - Actual: 5.0
Prediction mean: 6.75, stddev: 0.74, 95% CI: [8.19 - 5.3]  - Actual: 6.0
Prediction mean: 5.5,  stddev: 0.73, 95% CI: [6.93 - 4.07] - Actual: 5.0
```

compared to the result of a conventional neural network with three deterministic ('Dense') layers for the same samples:

```
Predicted: 5.8 - Actual: 6.0
Predicted: 5.7 - Actual: 5.0
Predicted: 5.9 - Actual: 6.0
Predicted: 6.3 - Actual: 6.0
Predicted: 6.3 - Actual: 8.0
Predicted: 5.8 - Actual: 5.0
Predicted: 4.9 - Actual: 6.0
Predicted: 5.1 - Actual: 5.0
Predicted: 6.4 - Actual: 6.0
Predicted: 5.8 - Actual: 5.0
```

The simple experiment shows that in TensorFlow Probability: (1) we can train a hybrid BNN as is, with both deterministic layers and probabilistic layers and the latter selectively positioned, and (2) positioning the probabilistic layer(s) at the end gives better result, consistent with the findings of other studies [12, 14-16].

## 7 Summary and Conclusion

Bayesian neural networks provide a direct and natural way to extend conventional deep neural networks to support probabilistic deep learning. However, it is nontrivial to understand, design and train Bayesian neural networks due to their complexities. We discussed the essentials of Bayesian neural networks including duality (deep neural networks, probabilistic models), approximate Bayesian inference, Bayesian priors, Bayesian posteriors, and deep variational learning.

The main problem with Bayesian neural networks is that the architecture of deep neural networks makes it quite redundant, and costly, to account for uncertainty for a large number of successive layers. Hybrid Bayesian neural networks, which use only few probabilistic layers likely positioned at the end of the networks, provide a practical solution.

Bayesian neural networks can improve the performance of conventional neural networks, both in terms of uncertainty estimation and prediction accuracy. With accurate and efficient deep variational learning algorithms and judiciously placed probabilistic layers, Bayesian neural networks are a viable approach to probabilistic deep learning.

**Acknowledgement:** Thanks to my wife Hedy (期芳) for her support.